\documentclass[10pt,twocolumn,letterpaper]{article}

\usepackage{iccv}
\usepackage{times}
\usepackage{epsfig}
\usepackage{graphicx}
\usepackage{amsmath}
\usepackage{amssymb}

\usepackage[ruled,vlined]{algorithm2e}

\usepackage{color}
\usepackage{comment}
\usepackage{enumitem}
\usepackage[export]{adjustbox}
\usepackage{tabularx}
\usepackage[table]{xcolor}
\usepackage{colortbl}
\usepackage{subfig}
\usepackage{multirow}
\usepackage{booktabs}
\usepackage{xr}
\usepackage{pifont}
\usepackage[hyphens]{url}
\usepackage{sidecap}
\usepackage{tablefootnote}

\usepackage[pagebackref=true,breaklinks=true,letterpaper=true,colorlinks,bookmarks=false]{hyperref}

\iccvfinalcopy 

\newcommand{\Paragraph}[1]{\vspace{1mm} \noindent \textbf{#1} \hspace{0mm}}

\newcommand{\cmark}{\ding{51}}%
\newcommand{\xmark}{\ding{55}}%
\newcommand{\camready}[1]{\textcolor{black}{#1}}

\def\iccvPaperID{3473} 

\begin{document}

\title{Mesh Graphormer}

\author{Kevin Lin \ \ \  Lijuan Wang \ \ \ Zicheng Liu\\
Microsoft\\
{\tt\small \{keli, lijuanw, zliu\}@microsoft.com}
}

\maketitle
\ificcvfinal\thispagestyle{empty}\fi

\begin{abstract}
We present a graph-convolution-reinforced transformer, named Mesh Graphormer, for 3D human pose and mesh reconstruction from a single image. Recently both transformers and graph convolutional neural networks (GCNNs) have shown promising progress in human mesh reconstruction. Transformer-based approaches are effective in modeling non-local interactions among 3D mesh vertices and body joints, whereas GCNNs are good at exploiting neighborhood vertex interactions based on a pre-specified mesh topology. In this paper, we study how to combine graph convolutions and self-attentions in a transformer to model both local and global interactions. Experimental results show that our proposed method, Mesh Graphormer, significantly outperforms the previous state-of-the-art methods on multiple benchmarks, including Human3.6M, 3DPW, and FreiHAND datasets. \camready{Code and pre-trained models are available at \url{https://github.com/microsoft/MeshGraphormer}.}

\end{abstract}

\section{Introduction}
3D human pose and mesh reconstruction from a single image is a popular research topic as it offers a wide range of applications for human-computer interactions. However, due to the complex body articulation, it is a challenging task.

Recently, Transformers and Graph Convolutional Neural Networks (GCNNs) have shown promising advances in human mesh reconstruction. For example, recent studies~\cite{kolotouros2019convolutional,Choi_2020_ECCV_Pose2Mesh} have suggested using GCNNs to directly regress 3D positions of mesh vertices by taking into account local interactions among neighbor vertices. In a recent study~\cite{lin2020end}, a transformer encoder was used with self-attention to capture global interactions among body joints and mesh vertices, which lead to further improvement.

\begin{figure}[h]
\begin{center}
\includegraphics[trim=0 0 0 0, clip,width=1.0\columnwidth]{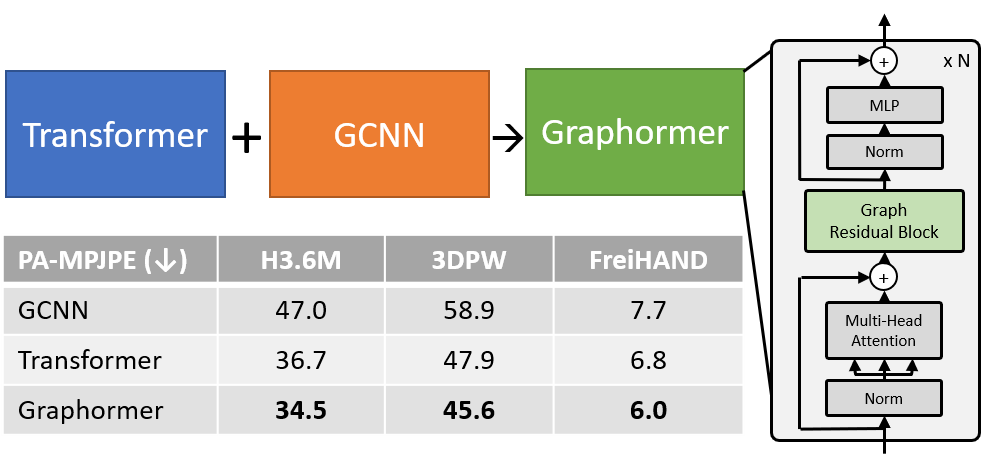}
\caption{
Summary. We study how to combine self-attentions and graph convolutions in a transformer for human mesh reconstruction. The proposed Graphormer outperforms existing graph convolution networks~\cite{Choi_2020_ECCV_Pose2Mesh,kolotouros2019convolutional} and transformer-based method~\cite{lin2020end} by a clear margin. The numbers are the reconstruction error (PA-MPJPE) in the unit of millimeter. The lower the better.
} 
\vspace{-6mm}
\label{fig:fig1}
\end{center}
\end{figure}

However, as discussed in the literature~\cite{yu2018qanet,wu2020lite,gulati2020conformer,yu2018qanet,chen20182}, Transformers and Convolution Neural Networks (CNN) each have their own limitations. Transformers are good at modeling long-range dependencies on the input tokens, but they are less efficient at capturing fine-grained local information. Convolution layers, on the other hand, are useful for extracting local features, but many layers are required to capture global context. In natural language processing and speech recognition, Conformer~\cite{gulati2020conformer} is a recently proposed technique that leverages the complementarity of self-attention and convolutions to learn representations. This motivates us to combine self-attention and graph convolution for the 3D reconstruction of human mesh (Figure~\ref{fig:fig1}). 

We present a graph-convolution-reinforced transformer called Mesh Graphformer for reconstructing human pose and mesh from a single image. We inject graph convolutions into transformer blocks to improve the local interactions among neighboring vertices and joints. In order to leverage the power of graph convolutions, Graphormer is free to attend 
to all image grid features that contain more detailed local information and are helpful in refining the 3D coordinate prediction. Consequently, Graphormer and image grid features are mutually reinforced to achieve better performance in human pose and mesh reconstruction.

Extensive experiments show that the proposed Mesh Graphormer models both local and global interactions effectively and clearly outperforms previous state-of-the-art methods for reconstructing human mesh in several datasets. Additionally, we offer ablation studies on various model design options to incorporate graph convolutions and self-attention into a transformer encoder.

The main contributions of this paper include 
\begin{itemize}
\item{We present a graph-convolution-reinforced transformer called Mesh Graphormer to model both local and global interactions for the 3D reconstruction of human pose and mesh.}
\item{Mesh Graphormer allows joints and mesh vertices to freely attend to image grid features to refine the 3D coordinate prediction.}
\item{Mesh Graphormer outperforms previous state-of-the-art methods on Human3.6M, 3DPW, and FreiHAND datasets.}
\end{itemize}
 
\section{Related Works}\label{sec:related}

\begin{figure}[t]
\begin{center}
\includegraphics[trim=0 0 0 0, clip,width=0.99\columnwidth]{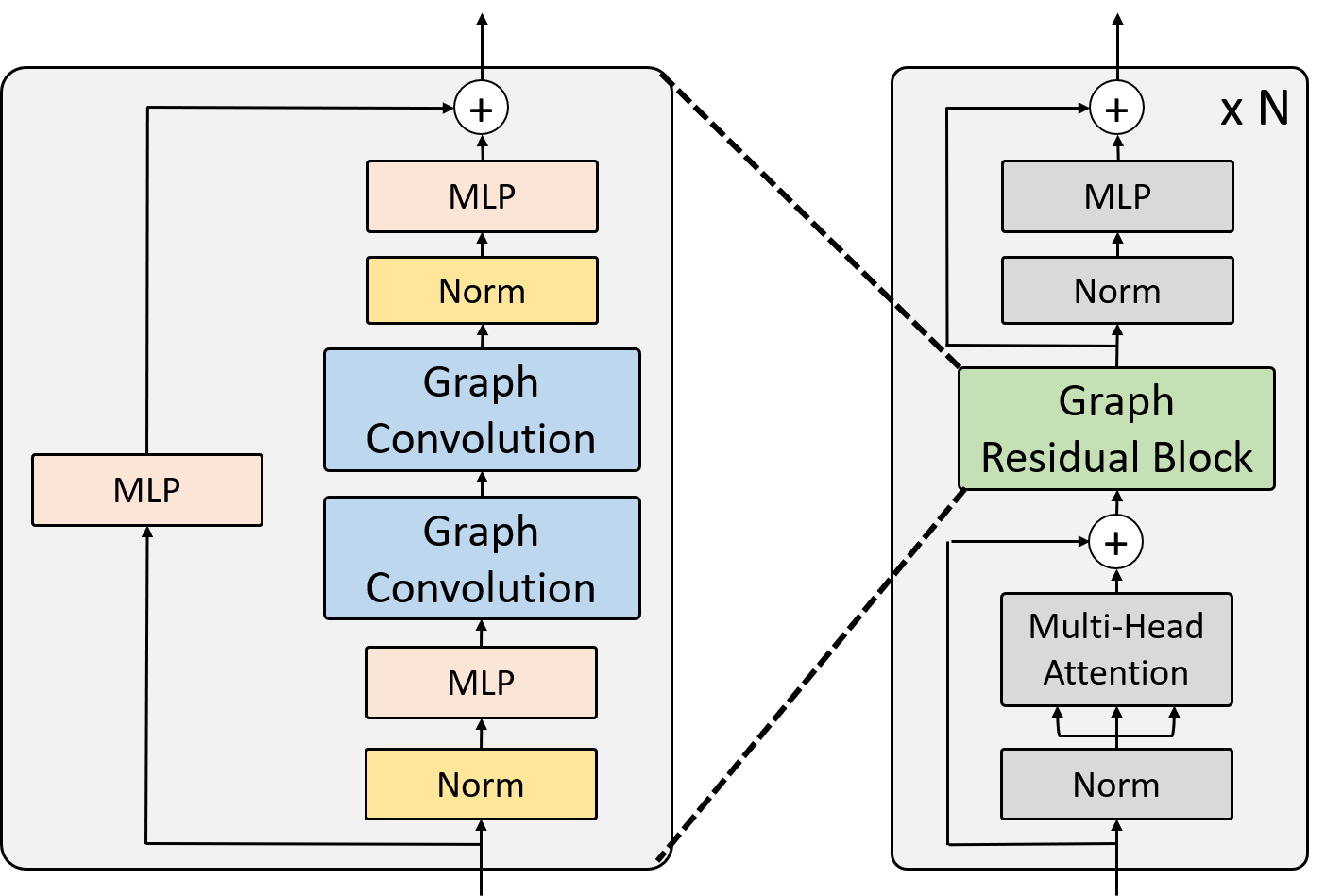}
\caption{
Architecture of a Graphormer Encoder. We propose a graph-convolution-reinforced transformer encoder to capture both global and local interactions for 3D human mesh reconstruction. The encoder consists of a stack of $N=4$ identical blocks.} 
\vspace{-4mm}
\label{fig:graphormer}
\end{center}
\end{figure}

\Paragraph{Human mesh reconstruction}can generally be divided into parametric and non-parametric approaches. The vast majority of previous work~\cite{kolotouros2019learning,kanazawa2018end,tung2017self,guan2009estimating,kocabas2019vibe,lassner2017unite,pavlakos2018learning} uses a parametric human model such as SMPL~\cite{loper2015smpl} and focuses on using the SMPL parameter space as a regression target. Given the pose and shape coefficients, SMPL is robust and practical for creating human mesh. However, as discussed in the literature~\cite{kolotouros2019convolutional,omran2018neural,Choi_2020_ECCV_Pose2Mesh, zhang2020learning}, it remains difficult to estimate accurate coefficients from a single image, and researchers~\cite{choi2020beyond, pavlakos2020human, kocabas2019vibe,moon2020pose2pose} are trying to predict 3D posture, learning with more visual evidence, or \camready{adopting the dense correspondence maps~\cite{zhang2020learning, guler2018densepose}} to improve the reconstruction.

Instead of regressing the parametric coefficients, nonparametric approaches~\cite {Choi_2020_ECCV_Pose2Mesh, Moon_2020_ECCV_I2L-MeshNet, kolotouros2019convolutional} regress vertices directly from an image. Among the previous studies, the Graph Convolutional Neural Network (GCNN)~\cite{kipf2016semi,Choi_2020_ECCV_Pose2Mesh, kolotouros2019convolutional} is one of the most popular options as it is able to model the local interactions between neighboring vertices based on a given adjacency matrix~\cite{ranjan2018generating,Choi_2020_ECCV_Pose2Mesh,kolotouros2019convolutional,wang2018pixel2mesh}. However, it is less efficient to capture global interactions between the vertices and body joints. To overcome this limitation, transformer-based methods~\cite{lin2020end} use a self-attention mechanism to freely attend vertices and body joints in the mesh and thereby encode non-local relationships of a human mesh. However, it is less convenient to model local interactions than GNN-based methods~\cite {Choi_2020_ECCV_Pose2Mesh, kolotouros2019convolutional}. 

Among the works mentioned above, METRO~\cite {lin2020end} is the most relevant study to our proposed method. The main difference between METRO and our proposed model is that we are designing a graph-convolution-reinforced transformer encoder for reconstructing human mesh. In addition, we add image grid features as input tokens to the transformer and allow the joints and mesh vertices to attend to grid features.

\Paragraph{Transformer architecture}is developing rapidly for different applications~\cite{devlin2019bert,vaswani2017attention,tay2020efficient,khan2021transformers,han2020survey}. One important direction is to improve the expressiveness of a transformer network for better context modeling. Recent studies~\cite {wu2020lite, gulati2020conformer} show that the combination of convolution and self-attention in a transformer encoder is helpful to improve representation learning. However, previous work has mainly focused on language modeling and speech recognition. When it comes to complex data structures such as 3D human mesh, this remains an open problem.

To address these challenges, we investigate how to inject graph convolutions into the transformer encoder blocks to better model both local and global interactions among 3D mesh vertices and body joints.

\begin{figure*}[t]
\begin{center}
\includegraphics[trim=0 0 0 0, clip,width=1\textwidth]{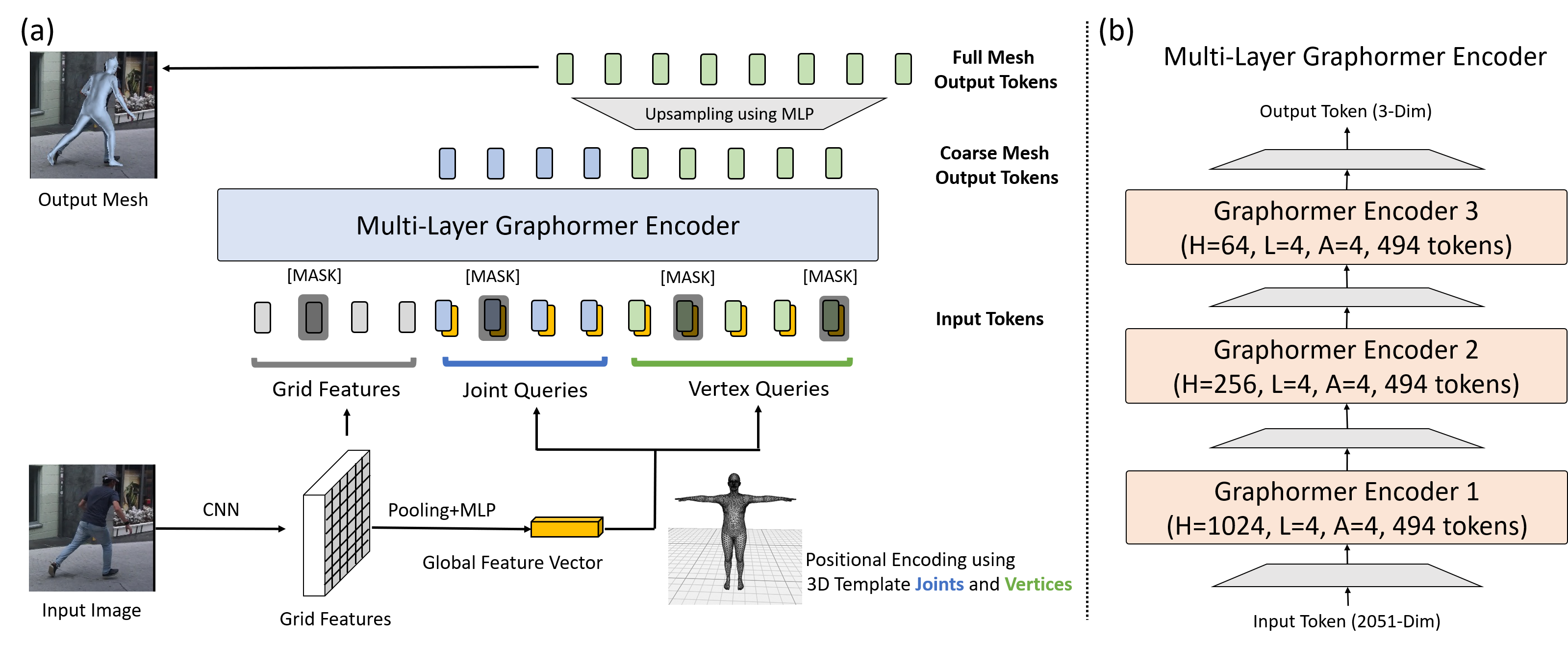}
\caption{
Graphormer for Human Mesh Reconstruction. (a) Our end-to-end mesh regression framework takes an image as input and predicts 3D joints and mesh vertices at the same time. We extract both the grid features and a global feature vector using a pre-trained CNN. The features are then tokenized and fed to a multi-layer Graphormer encoder. Grid features and multi-layer Graphormer encoder are mutually reinforced to reconstruct an accurate human mesh. (b) Our multi-layer Graphormer encoder consists of three Graphormer encoder blocks. All encoders have the same number of input tokens. We gradually reduce the hidden sizes, and output the 3D coordinates of body joints and mesh vertices.
} 
\vspace{-4mm}
\label{fig:overview}
\end{center}
\end{figure*}

\section{Graphormer Encoder}\label{sec:graphormer}

Figure~\ref{fig:graphormer} shows the architecture of our proposed Graphormer encoder. It consists of a stack of $N=4$ identical blocks. Each block consists of five sub-modules, including a Layer Norm, a Multi-Head Self-Attention module, a Graph Residual Block, a second Layer Norm, and a Multi-Layer Perceptron (MLP) at the end. Our Graphormer encoder has a similar architecture to the traditional transformer encoder~\cite {vaswani2017attention}, but we introduce graph convolution into the network to model fine-grained local interactions. In the following sections, we describe Multi-Head Self-Attention (MHSA) module and Graph Residual Block.

\subsection{Multi-Head Self-Attention}

We employ the Multi-Head Self-Attention (MHSA) module proposed by Vaswani~\etal~\cite{vaswani2017attention}, which uses several self-attention functions in parallel to learn contextual representation. Given an input sequence ${\bf{X}} = \{{\bf{x}}_1, {\bf{x}}_2, \dots, {\bf{x}}_n\} \in \mathbb{R}^{n\times d}$, where $d$ is the hidden size. It first projects the input sequence to queries $\bf{Q}$, keys $\bf{K}$, and values $\bf{V}$ by using trainable parameters $\{ {\bf{W}}_Q, {\bf{W}}_K, {\bf{W}}_V\} \in \mathbb{R}^{d\times d}$. That is written as \begin{equation}
\label{eqn:attention}%
{\bf{Q}},{\bf{K}},{\bf{V}} = {\bf{XW}}_Q, {\bf{XW}}_K, {\bf{XW}}_V \ \ \ \ \ \in \mathbb{R}^{n\times d}.
\end{equation}

The three feature representations are split into $h$ different subspaces, e.g., ${\bf{Q}} = [{\bf{Q}}^1, {\bf{Q}}^2, \dots, {\bf{Q}}^h]$ where ${\bf{Q}}^i \in \mathbb{R}^{n\times \frac{d}{h}}$, so that we can perform self-attention for each subspace individually. For each subspace, the output ${\bf{Y}}^h = \{ {\bf{y}}_1^h, {\bf{y}}_2^h, \dots, {\bf{y}}_n^h\}$ are computed by:\begin{equation}
\label{eqn:attention}%
{\bf{y}}_i^h = {\bf{Att}}({\bf{q}}_i^h, {\bf{K}}^h){\bf{V}}^h \ \ \ \ \ \in \mathbb{R}^{\frac{d}{h}},
\end{equation}
where ${\bf{Att}(\cdot)}$ denotes the attention function~\cite{vaswani2017attention} that quantifies how semantically relevant a query ${\bf{q}}_i^h$ is to keys ${\bf{K}}^h$ by scaled dot-product and softmax. The output ${\bf{Y}}^h \in \mathbb{R}^{n\times \frac{d}{h}}$ from each subsapce are later concatenated to form the final output ${\bf{Y}} \in \mathbb{R}^{n\times d}$.

\subsection{Graph Residual Block}

While MHSA is useful for extracting long-range dependencies, it is less efficient at capturing fine-grained local information in complex data structures such as 3D mesh. The Graph Residual Block in our proposed network aims to address this challenge. 

Given the contextualized features ${\bf{Y}} \in \mathbb{R}^{n\times d}$ generated by MHSA, we improve the local interactions with the help of graph convolution:\begin{equation}
\label{eqn:abstract}
{\bf{Y'}} = {\bf{GraphConv}}({\bf{\bar{A}}},{\bf{Y}};{\bf{W}}_G) = \sigma({\bf{\bar{A}}}{\bf{Y}}{\bf{W}}_G).
\end{equation}
${\bf{\bar{A}}} \in \mathbb{R}^{n\times n}$ denotes the adjacency matrix of a graph and ${\bf{W}}_G$ the trainable parameters.
$\sigma(\cdot)$ is the activation function that gives the network non-linearity. According to BERT~\cite{devlin2019bert}, we use the Gaussian Error Linear Unit (GeLU)~\cite{hendrycks2016gaussian} in this work.

To train with multiple graph convolution layers, we follow the design principle in GraphCMR~\cite{kolotouros2019convolutional} to build our Graph Residual Block. The network architecture is spiritually similar to~\cite{kolotouros2019convolutional}, but we replace the group normalization~\cite{wu2018group} with the layer normalization~\cite{ba2016layer} and replace ReLU~\cite{nair2010rectified} with GeLU~\cite{hendrycks2016gaussian}. This is to bring the type of layers in line with the transformer.

Our Graph Residual Block makes it possible to explicitly encode the graph structure within the network and thereby improve spatial locality in the features.

\section{Graphormer for Mesh Reconstruction}\label{sec:HMR}

In this section, we describe how the Graphormer encoder is applied to human pose and mesh reconstruction from a single image.

Figure~\ref{fig:overview}(a) gives an overview of our end-to-end mesh regression framework. An image with a size of 224x224 is used as input and image grid features are extracted. 
The image features are tokenized as input for a multi-layer Graphormer encoder. In the end, our end-to-end framework predicts 3D coordinates of the mesh vertices and body joints at the same time.

In the following, we describe the image-based architecture with which we extract image grid features. Next, we describe the Multi-Layer Graphormer Encoder for regression of the 3D vertices and body joints. Finally, we discuss important training details.

\subsection{CNN and Image Grid Features}

In the first part of our model, we use a pre-trained image-based CNN for feature extraction. Previous work~\cite{kolotouros2019convolutional, kanazawa2018end, lin2020end} extracts a global 2048-Dim image feature vector as a model input. The disadvantage is that a global feature vector does not contain fine-grained local details. This motivates us to add grid features~\cite{jiang2020defense} as input tokens and allow joints and mesh vertices to freely attend to all the grid features. As we will show in our experiments, the local information provided by the grid features are effectively leveraged by the graph convolutions in the Graphormer to refine 3D positions of mesh vertices and body joints.

As shown in Figure~\ref{fig:overview}(a), we extract the grid features from the last convolution block in CNN. The grid features are typically $7 \times7 \times1024$ in size. They are tokenized to $49$ tokens, and each token is a $1024$-Dim vector. Similar to~\cite {lin2020end}, we also extract the $2048$-Dim image feature vector from the last hidden layer of the CNN and perform positional encoding using the 3D coordinates of each vertex and body joint in a human template mesh. Finally, we apply MLP to make the size of all input tokens consistent. After that, all input tokens have $2051$-Dim.

\subsection{Multi-Layer Graphormer Encoder}

Given the grid features, joint queries, and vertex queries, our multi-layer Graphormer encoder sequentially reduces the dimensions to map the inputs to 3D body joints and mesh vertices at the same time.

As shown in Figure~\ref{fig:overview}(b), our multi-layer Graphormer encoder consists of three encoder blocks. The three encoder blocks have the same number of tokens including $49$ grid feature tokens, $14$ joint queries, and $431$ vertex queries. However, the three encoders have different hidden dimensions. In this work, the hidden dimensions of the three encoders are $1024$, $256$ and $64$, respectively.

Similar to~\cite{lin2020end}, our multi-layer Graphormer encoder processes a coarse mesh with 431 vertices. We use a coarse template mesh for positional encoding, and our Graphormer encoder outputs a coarse mesh. Then, we use a linear projection to sample the coarse mesh up to the original resolution (with 6K vertices in SMPL mesh topology). As explained in the literature~\cite {kolotouros2019convolutional,lin2020end}, learning a coarse mesh followed by upsampling is helpful to avoid redundancies in the original mesh and makes training more efficient.

\subsection{Training Details}

Following METRO~\cite{lin2020end}, we train our model by applying a loss of $ L_1 $ in addition to the Graphormer outputs. In our training, we also use Masked Vertex Modeling~\cite{lin2020end} to improve the robustness of our model. To be specific, we apply $L_1$ loss to 3D mesh vertices and body joints. We also apply $L_1$ loss to 2D projected body joints to improve the alignment between the image and the reconstructed mesh. In addition, we apply intermediate supervision on the coarse mesh to accelerate convergence.

We train Graphormers with the Adam optimizer~\cite{kingma2014adam}. We use an initial learning rate of $1\times10^{-4}$ for both Graphormer and CNN backbones. We train our model for $200$ epochs and lower the learning rate by a factor of $10$ after $100$ epochs. All Graphormer weights are randomly initialized, and the CNN backbone is initialized with ImageNet pre-trained weights. Following~\cite{lin2020end}, we report on results with the HRNet backbone~\cite{WangSCJDZLMTWLX19}.

\begin{figure*}
\begin{center}
\includegraphics[trim=0 0 0 0, clip,width=0.33\textwidth]{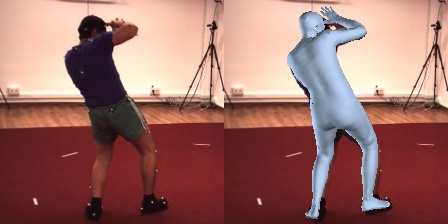}
\includegraphics[trim=0 0 0 0, clip,width=0.33\textwidth]{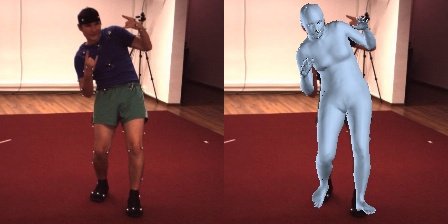}
\includegraphics[trim=0 0 0 0, clip,width=0.33\textwidth]{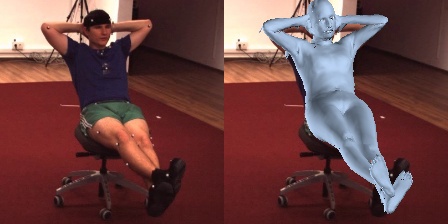}\\
\includegraphics[trim=0 0 0 0, clip,width=0.33\textwidth]{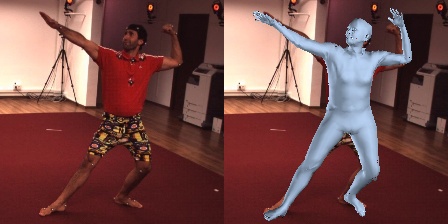}
\includegraphics[trim=0 0 0 0, clip,width=0.33\textwidth]{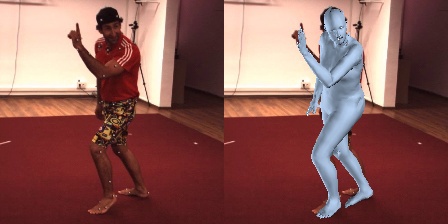}
\includegraphics[trim=0 0 0 0, clip,width=0.33\textwidth]{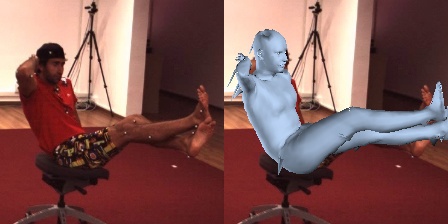}\\
\includegraphics[trim=0 0 0 0, clip,width=0.33\textwidth]{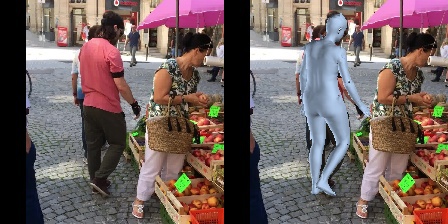}
\includegraphics[trim=0 0 0 0, clip,width=0.33\textwidth]{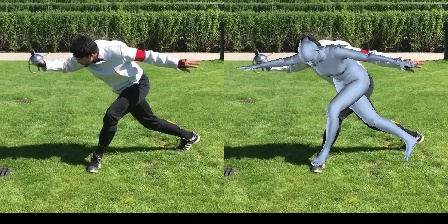}
\includegraphics[trim=0 0 0 0, clip,width=0.33\textwidth]{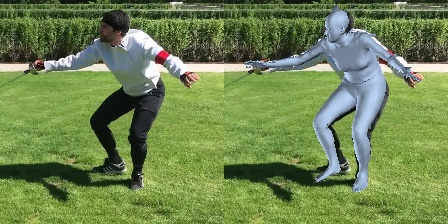}\\
\includegraphics[trim=0 0 0 0, clip,width=0.33\textwidth]{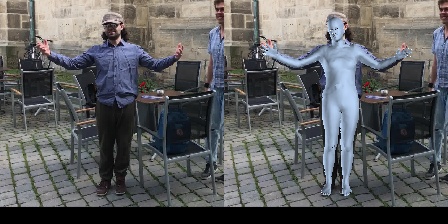}
\includegraphics[trim=0 0 0 0, clip,width=0.33\textwidth]{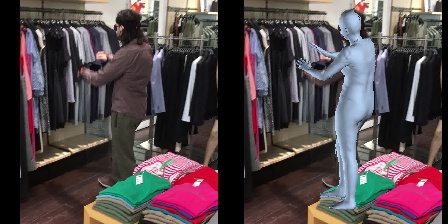}
\includegraphics[trim=0 0 0 0, clip,width=0.33\textwidth]{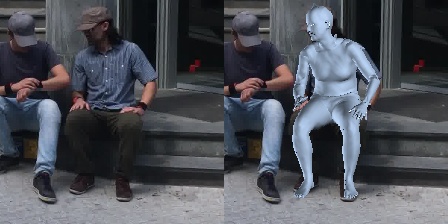}\\

\caption{
Qualitative results of our method on different datasets. The top two rows are the results on Human3.6M, and the bottom two rows are the results on 3DPW.
} 
\vspace{-2mm}
\label{fig:vis_h36m}
\end{center}
\end{figure*} 

\begin{table*}[t]
\centering
\begin{tabular}{lcccccc}
    \toprule
    \multirow{1}{*}{} & \multicolumn{3}{c}{3DPW} & & \multicolumn{2}{c}{Human3.6M}\\ 
    \cline{2-4}\cline{6-7}
	Method  & MPVE $\downarrow$ & MPJPE $\downarrow$ & PA-MPJPE $\downarrow$ & &  MPJPE $\downarrow$ & PA-MPJPE $\downarrow$ \\
	\midrule
	HMR~\cite{kanazawa2018end} & $-$ & $-$ & $81.3$ && $88.0$ & $56.8$  \\
	GraphCMR~\cite{kolotouros2019convolutional} & $-$ & $-$ & $70.2$ && $-$ & $50.1$\\
	SPIN~\cite{kolotouros2019learning} & $116.4$ & $-$ & $59.2$ && $-$ & $41.1$\\
	Pose2Mesh~\cite{Choi_2020_ECCV_Pose2Mesh} & $-$ & $89.2$ & $58.9$ && $64.9$ & $47.0$\\
	I2LMeshNet~\cite{Moon_2020_ECCV_I2L-MeshNet} & $-$ & $93.2$ & $57.7$ && $55.7$ & $41.1$\\
	VIBE~\cite{kocabas2019vibe} & $99.1$ & $82.0$ & $51.9$ && $65.6$ & $41.4$\\
    METRO~\cite{lin2020end}  & $88.2$ & $77.1$ & $47.9$ && $54.0$ & $36.7$\\
    \midrule
    Mesh Graphormer  & $\textbf{87.7}$ & $\textbf{74.7}$ & $\textbf{45.6}$ && $\textbf{51.2}$ & $\textbf{34.5}$\\
	\bottomrule
\end{tabular}
\caption{Performance comparison with the previous state-of-the-art methods on 3DPW and Human3.6M datasets. 
}
\vspace{-2mm}
\label{tbl:compare-h36m-3dpw}
\end{table*}

\begin{figure*}
\begin{center}
\includegraphics[trim=0 0 0 0, clip,width=0.33\textwidth]{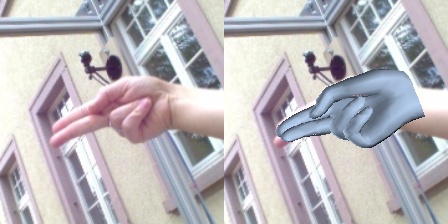}
\includegraphics[trim=0 0 0 0, clip,width=0.33\textwidth]{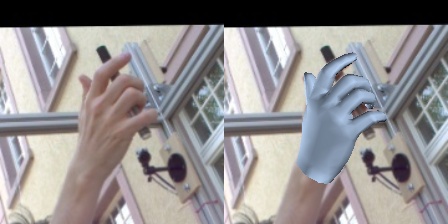}
\includegraphics[trim=0 0 0 0, clip,width=0.33\textwidth]{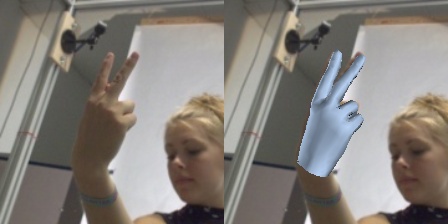}\\
\includegraphics[trim=0 0 0 0, clip,width=0.33\textwidth]{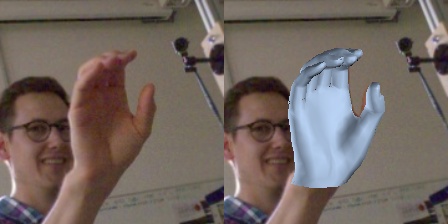}
\includegraphics[trim=0 0 0 0, clip,width=0.33\textwidth]{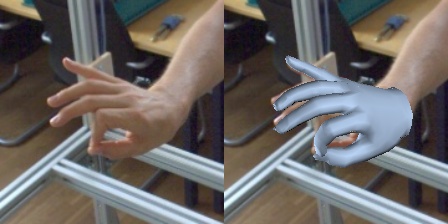}
\includegraphics[trim=0 0 0 0, clip,width=0.33\textwidth]{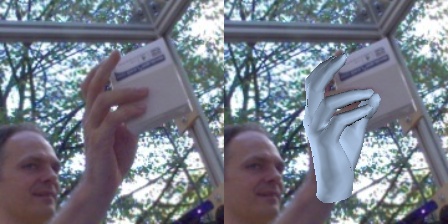}\\
\includegraphics[trim=0 0 0 0, clip,width=0.33\textwidth]{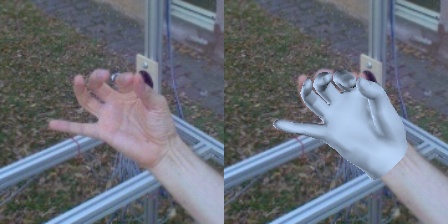}
\includegraphics[trim=0 0 0 0, clip,width=0.33\textwidth]{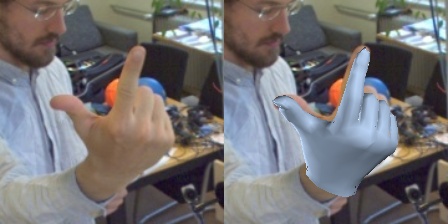}
\includegraphics[trim=0 0 0 0, clip,width=0.33\textwidth]{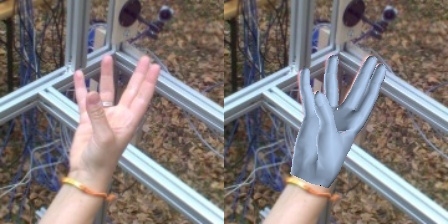}\\
\includegraphics[trim=0 0 0 0, clip,width=0.33\textwidth]{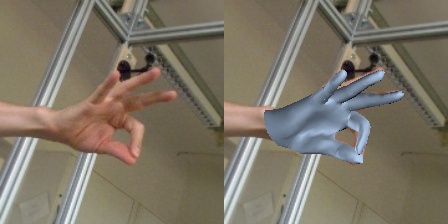}
\includegraphics[trim=0 0 0 0, clip,width=0.33\textwidth]{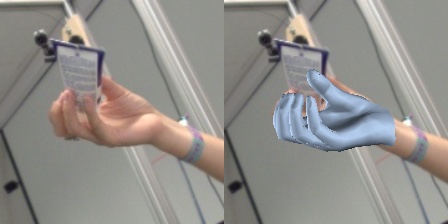}
\includegraphics[trim=0 0 0 0, clip,width=0.33\textwidth]{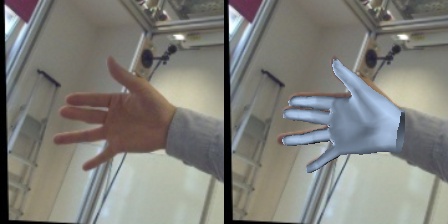}\\
\caption{
Qualitative results of our method on FrieHAND test set. Graphormer generalizes to 3D hand reconstruction even if the gestures are novel and complicated.
} 
\label{fig:vis_freihand}
\vspace{-2mm}
\end{center}
\end{figure*} 

\begin{table*}[t]
\centering
\begin{tabular}{lcccc}
    \toprule
	Method  & PA-MPVPE $\downarrow$ & PA-MPJPE $\downarrow$ & F@5 mm $\uparrow$ & F@15 mm $\uparrow$\\
	\midrule
	Hasson et al~\cite{kanazawa2018end} & $13.2$ & $-$ & $0.436$ & $0.908$\\
	Boukhayma et al.~\cite{kolotouros2019convolutional} & $13.0$ & $-$ & $0.435$ & $0.898$\\
	FreiHAND~\cite{kolotouros2019learning} & $10.7$ & $-$ & $0.529$ & $0.935$\\
	Pose2Mesh~\cite{Choi_2020_ECCV_Pose2Mesh} & $7.8$ & $7.7$ & $0.674$ & $0.969$\\
	I2LMeshNet~\cite{Moon_2020_ECCV_I2L-MeshNet} & $7.6$ & $7.4$ & $0.681$ & $0.973$\\
    METRO~\cite{lin2020end}  & $6.7$ & $6.8$ & $0.717$ & $0.981$\\
    \midrule
    Mesh Graphormer   & $\textbf{5.9}$ & $\textbf{6.0}$ & $\textbf{0.764}$ & $\textbf{0.986}$\\
	\bottomrule
\end{tabular}
\caption{Performance comparison with the previous state-of-the-art methods on FreiHAND dataset. }
\label{tbl:compare-hand}
\vspace{-2mm}
\end{table*}

\section{Experiments}\label{sec:exp}
In this section, we first discuss the datasets we use in our training. We then compare our method with previous state-of-the-art methods. Finally, we provide in-depth ablation studies on our model architectures.

\subsection{Datasets}
We conduct extensive training using publicly available datasets, including Human3.6M~\cite{ionescu2014human3}, MuCo-3DHP~\cite{mehta2018single}, UP-3D~\cite{lassner2017unite},  COCO~\cite{lin2014microsoft}, MPII~\cite{andriluka14cvpr}. Please note that the 3D mesh training data from Human3.6M~\cite{ionescu2014human3} is not available due to licensing issues. Therefore, we use the pseudo 3D mesh training data from~\cite{Choi_2020_ECCV_Pose2Mesh,Moon_2020_ECCV_I2L-MeshNet}. We follow the general setting where subjects S1, S5, S6, S7 and S8 are used in training, and test subjects S9 and S11. We present all results using the P2 protocol~\cite{kanazawa2018end,kolotouros2019learning}. To get a fair comparison with earlier state-of-the-art~\cite{kocabas2019vibe,lin2020end}, we also use 3DPW~\cite{vonMarcard2018} training data for 3DPW experiments.

We also apply our method to reconstructing 3D hand. We conduct training on FreiHAND~\cite{zimmermann2019freihand} and evaluate using its online server. Please note that we use test-time augmentation for FreiHAND experiment.

\subsection{Main Results}

We compare our method with previous approaches to reconstruct human mesh on Human3.6M and 3DPW datasets. In Table~\ref{tbl:compare-h36m-3dpw}, we can see that our method outperforms the previous state-of-the-art for both datasets. This includes both GCNN-based models~\cite{kolotouros2019convolutional,Choi_2020_ECCV_Pose2Mesh} and transformer-based methods~\cite{lin2020end}. It suggests that Graphormer models both local and global interactions better and achieves more accurate mesh reconstruction. 

Figure~\ref{fig:vis_h36m} shows the qualitative results of our method on Human3.6M and 3DPW datasets. We see that our method is robust to challenging poses and noisy backgrounds. 

We also apply our method to 3D hand reconstruction and compare it with other modern approaches~\cite{lin2020end,Choi_2020_ECCV_Pose2Mesh,Moon_2020_ECCV_I2L-MeshNet,zimmermann2019freihand} on FreiHAND dataset. In Table~\ref{tbl:compare-hand}, we see that our method works much better than the prior art. This demonstrates the generalization capability of our method for other objects. 

Figure~\ref{fig:vis_freihand} shows the qualitative results of our method on FreiHAND test set. We can see that Graphormer can be used to reconstruct a 3D hand even if the target gesture is novel and complicated.

\subsection{Ablation Study}

We conduct a large-scale ablation study on Human3.6M~\cite{ionescu2014human3} to investigate our model capability. We evaluate our model using the 3D pose accuracy, and report the accuracy using PA-MPJPE metric~\cite{zhou2018monocap,kanazawa2018end,kolotouros2019learning,kolotouros2019convolutional}.

\Paragraph{Effectiveness of Grid Features:} In our first ablation study, we are interested in the effect of adding image grid features to the transformer encoder. We have implemented a baseline model~\cite{lin2020end} that uses the image grid features as inputs. Table~\ref{tbl:abalation-01} shows that the addition of image grid features improves the reconstruction performance by a clear margin. This indicates that using a single global feature vector is one of the performance bottlenecks in existing techniques.

\begin{table}
\centering
	\vspace{0mm}
\begin{tabular}{lc}
    \toprule
	Method  & PA-MPJPE\\
	\midrule
	Transformer~\cite{lin2020end}  & $36.7$\\
	Transformer + Grid Features  & $\textbf{35.9}$\\
	\bottomrule
\end{tabular}
\vspace{0mm}
\caption{Analysis of the effective of using the image grid features in a transformer framework.}
\label{tbl:abalation-01}
\end{table}

\begin{table}
\centering
	\vspace{0mm}
\begin{tabular}{lcccccc}
    \toprule
	Grid Features  & A & B & C & D & E & F\\
	\midrule
	PA-MPJPE  & $37.1$ & $37.7$ & $36.3$ & $38.1$ & $36.0$ & $\textbf{35.9}$\\
	\bottomrule
\end{tabular}
\vspace{0mm}
\caption{Performance comparison of the use of different image grid features from the HRNet.}
\label{tbl:abalation-grid}
\end{table} 

\begin{figure}[t]
\begin{center}
\includegraphics[trim=0 0 0 0, clip,width=1\columnwidth]{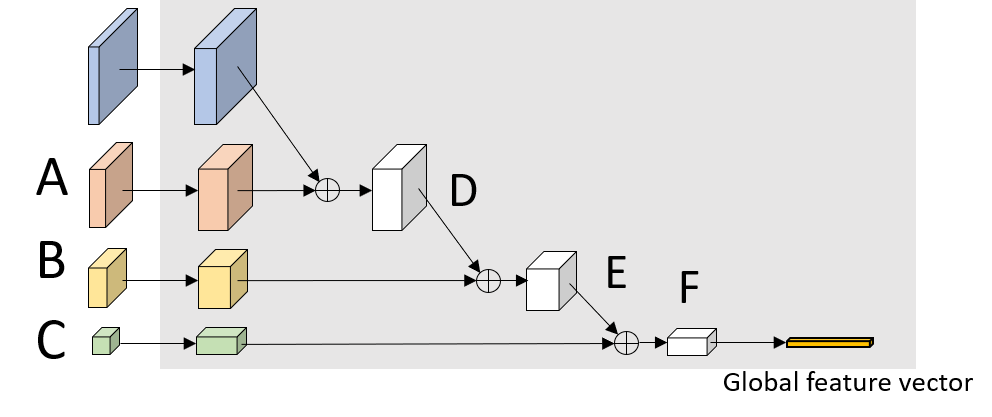}
\caption{
Illustration of different grid features we extracted from HRNet~\cite{WangSCJDZLMTWLX19} for ablation study. 
} 
\vspace{-0mm}
\label{fig:pyramid}
\end{center}
\end{figure}

The next question one may ask is what feature maps are important for training a transformer encoder. We perform analysis for a variety of feature maps in an HRNet~\cite{WangSCJDZLMTWLX19}. Figure~\ref{fig:pyramid} shows the six feature maps we have selected for experiments, and Table~\ref{tbl:abalation-grid} shows the performance comparison. Our finding is that the latest feature map (labeled F in Figure~\ref{fig:pyramid}) works better than others. The results suggest that the feature pyramid is helpful for improving the performance of a transformer encoder.

\begin{table}
\centering
	\vspace{0mm}
\begin{tabular}{cccc}
    \toprule
	Encoder1 & Encoder2 & Encoder3 & PA-MPJPE\\
	\midrule
	\xmark  & \xmark & \xmark & $35.9$\\
	\midrule
	\cmark  & \xmark & \xmark & $36.7$\\
	\xmark  & \cmark & \xmark & $36.2$\\
	\xmark  & \xmark & \cmark & $\textbf{35.1}$\\
	\cmark  & \cmark & \cmark & $36.0$\\
	\bottomrule
\end{tabular}
\vspace{0mm}
\caption{Ablation study of Graphormer Block for three encoders used for human mesh regression. \cmark: Add graph convolution into the specified transformer encoder; \xmark: No graph convolution in the specified transformer encoder.}
\label{tbl:abalation-which}
\end{table}

\Paragraph{Effectiveness of Adding Graph Convolution:} An important question is whether adding a graph convolution to the transformer is useful. In this experiment, we investigate the effect of graph convolution and how graph convolution and transformer can be combined.

Since our proposed framework contains three encoder blocks, we examine the effect of graph convolution by gradually adding a graph convolution layer to each encoder and comparing the performance. In Table~\ref{tbl:abalation-which}, the first row corresponds to the baseline that does not use any graph convolution in the network. The rest of the rows show the results of adding a graph convolution layer to different encoder blocks. The results show some interesting observations: (\textit{i}) Adding a graph convolution to Encoder1 or Encoder2 does not improve performance. (\textit{ii}) Adding the graph convolution to Encoder3 improves $ 0.9 $ PA-MPJPE. The results suggest that the lower layers focus on the global interactions of the mesh vertices to model the posture of the human body, while the upper layers pay more attention to the local interactions for better shape reconstruction.

\begin{table}
\centering
	\vspace{0mm}
\begin{tabular}{lc}
    \toprule
	Model Architecture  & PA-MPJPE\\
	\midrule
	Graph Conv. and MHSA in parallel & $36.4$\\
	Graph Conv. before MHSA & $35.6$\\
	Graph Conv. after MHSA & $\textbf{35.1}$\\
	\bottomrule
\end{tabular}
\vspace{0mm}
\caption{Performance comparison between different design options of the architecture of Graphormer Block.}
\label{tbl:abalation-graph}
\end{table} 

\Paragraph{Analysis of Encoder Architecture:} We further examine the network architecture in an encoder block. That is, we investigate the relationship between the MHSA module (Multi-Head Self-Attention) and the Graph Convolution module by using three different designs: (\textit{i}) We use a graph convolution layer and MHSA in parallel, similar to~\cite{wu2020lite}. (\textit{ii}) We first use a graph convolution layer and then MHSA. (\textit{iii}) We first use MHSA and then a graph convolution layer. \camready{Please refer to the supplementary material for a graphical illustration of the three architecture designs.} In Table~\ref{tbl:abalation-graph}, we can see that adding a graph convolution layer directly after MHSA works much better than other design options.

\begin{table}
\centering
	\vspace{0mm}
\begin{tabular}{lc}
    \toprule
	Design Choices & PA-MPJPE\\
	\midrule
	Grid Features & $35.9$\\
	Grid Features + Graph Conv. & $35.1$\\
	Grid Features + Graph Res. Block & $\textbf{34.5}$\\
	\bottomrule
\end{tabular}
\vspace{0mm}
\caption{Ablation study of the basic graph convolution and Graph Residual Block.}
\label{tbl:abalation-grb}
\end{table} 

\begin{figure*}[t]
\begin{center}
\includegraphics[trim=0 0 0 0, clip,width=1\textwidth]{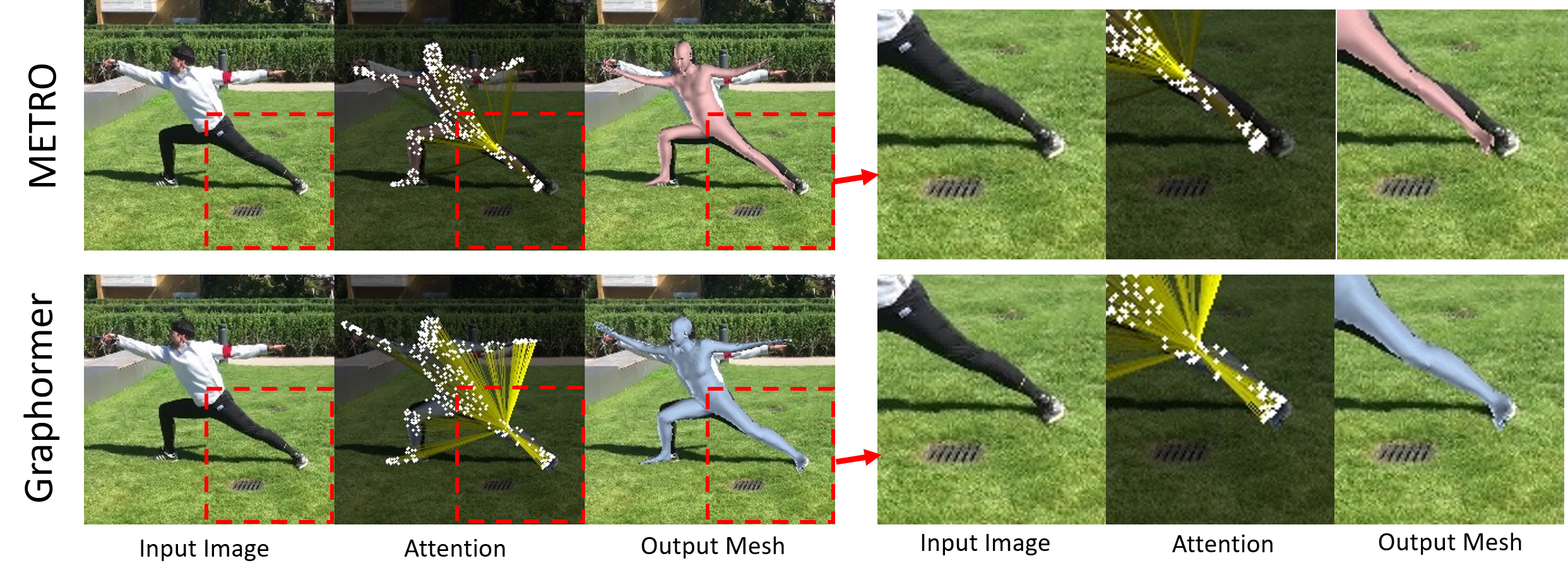}
\vspace{-8mm}
\caption{
Qualitative comparison. The top row (with pink mesh) shows the results of a transformer baseline~\cite{lin2020end} which has no graph convolutions. The bottom row (with blue mesh) shows our results with Graphormer. The attention visualization shows the interactions between the left knee and all other vertices, and the brighter color indicates stronger interactions. As can be seen, Graphormer is helpful to model both global and local interactions among vertices and body joints, and achieves a more accurate reconstruction. \camready{Please note the color of each attention map is normalized with respect to the maximum attention. Figure 13 in the supplementary material shows the attention maps without color normalization.}} 
\vspace{-4mm}
\label{fig:attention}
\end{center}
\end{figure*}

\Paragraph{Effectiveness of Adding Graph Residual Block:} In the previous sections, we explored the use of a basic graph convolution layer in a transformer encoder. Here we extend our experiments to Graph Residual Block. In particular, we replace a graph convolution layer with a Graph Residual Block to improve model capacity. In Table~\ref{tbl:abalation-grb}, we see that a Graph Residual Block works better than a basic graph convolution layer.

\begin{table}
\centering
	\vspace{0mm}
\begin{tabular}{cccc}
    \toprule
	Grid Features  & Graph Conv. & Graph Res. & PA-MPJPE\\
	\midrule
	\xmark  & \xmark & \xmark & $36.7$\\
	\xmark  & \cmark & \xmark & $36.7$\\
	\xmark  & \cmark & \cmark & $36.6$\\
	\midrule
	\cmark  & \xmark & \xmark & $35.9$\\
	\cmark  & \cmark & \xmark & $35.1$\\
	\cmark  & \cmark & \cmark & $\textbf{34.5}$\\
	\bottomrule
\end{tabular}
\vspace{-2mm}
\caption{\textcolor{black}{Ablation study of different combinations of image grid features and graph convolutions.}}
\label{tbl:abalation-grid-grb}
\end{table}

\Paragraph{Relationship between Grid Features and Graph Convolution:} An interesting question is what happens when we use Graphormer Encoder, but not grid features. To answer the question, Table~\ref{tbl:abalation-grid-grb} shows the comparisons. 

First of all, the first row of Table~\ref{tbl:abalation-grid-grb} corresponds to the baseline which has no grid features or graph convolution in the transformer. Compared to the fourth row of Table~\ref{tbl:abalation-grid-grb}, we see that we can achieve an improvement of $0.8$ PA-MPJPE if we only enable grid features. Next, if we enable graph convolution alone, as shown in the third row of Table~\ref{tbl:abalation-grid-grb}, only $0.1$ PA-MPJPE is improved. Finally, as shown in the bottom row of Table~\ref{tbl:abalation-grid-grb}, when we enable both grid features and graph convolution, it eventually improves PA-MPJPE by $2.2$ which is much greater than the sum of the two individual improvements ($0.1+0.8$). It shows that grid features and graph convolutions mutually reinforce each other, which leads to an further increase in performance.

\Paragraph{Visualization of Local Interactions:} We further study the effect of Graphormer in learning both global and local interactions among body joints and mesh vertices. We extract the attention maps from the last layer of our last encoder (i.e., Encoder3), and compute the average attentions of all the attention heads. We compare the self-attentions of Graphormer with that of the existing approach~\cite{lin2020end}. Figure~\ref{fig:attention} shows the qualitative comparison. We can see in the top row of Figure~\ref{fig:attention}, previous approach~\cite{lin2020end} fails to model the interactions between the left knee and left toes. In contrast, as shown in the bottom row of Figure~\ref{fig:attention}, Graphormer is able to model both global and local interactions, especially those between the left knee and the left toes. As a result, Graphormer reconstructs a more favorable shape compared to prior works.

\section{Conclusion}\label{sec:conclusion}

We introduced Mesh Graphormer, a new transformer architecture that incorporates graph convolutions and self-attentions for reconstructing human pose and mesh from a single image. We explored various model design options and demonstrated that both graph convolutions and grid features are helpful for improving the performance of the transformer. Experimental results show that our method generates new state-of-the-art performance on Human3.6M, 3DPW, and FreiHAND datasets.

{\small
\bibliographystyle{ieee_fullname}
\bibliography{humanmesh}
}

\clearpage
\appendix
\pdfoutput=1

\twocolumn[{%
\renewcommand\twocolumn[1][]{#1}%
\begin{center}
\textbf{\Large Supplementary Material}
\end{center}
\vspace{2mm}
\begin{center}
	\centering
\includegraphics[trim=0 0 0 0, clip,width=0.49\textwidth]{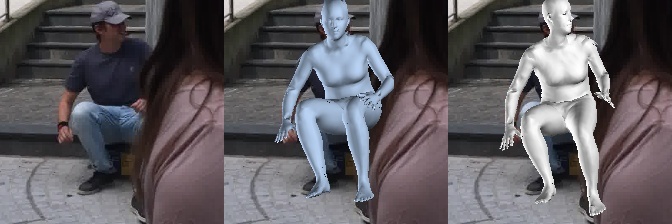}
\includegraphics[trim=0 0 0 0, clip,width=0.49\textwidth]{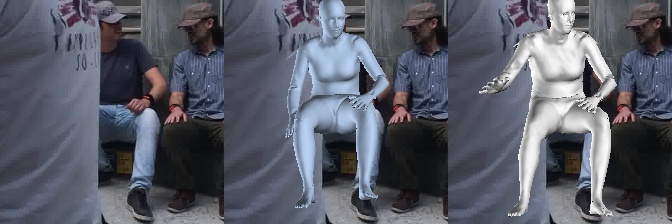}\\
\includegraphics[trim=0 0 0 0, clip,width=0.49\textwidth]{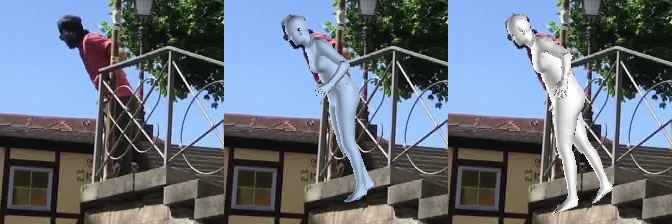}
\includegraphics[trim=0 0 0 0, clip,width=0.49\textwidth]{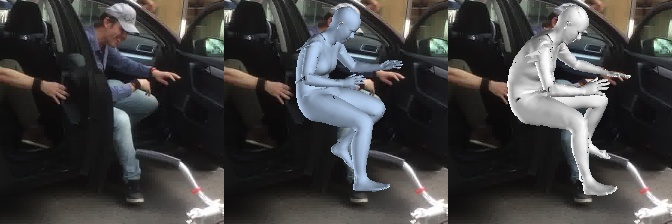}
\setlength{\tabcolsep}{25pt}
\begin{tabular}{cccccc}
Input & METRO &  Graphormer & Input &  METRO & Graphormer
\end{tabular}
\captionof{figure}{
	Qualitative results when there are heavy occlusions. For each example, we show results from METRO~\cite{lin2020end} and Graphormer. We can see that both METRO and Graphormer are quite robust against occlusions, but Graphormer generates more favorable head pose and body pose. Blue: METRO. Silver: Graphormer.}
	\label{fig:occlusions}
\end{center}%
}]

\begin{figure*}
\begin{center}
\includegraphics[trim=0 0 0 0, clip,width=0.33\textwidth]{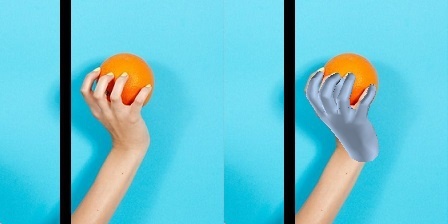}
\includegraphics[trim=0 0 0 0, clip,width=0.33\textwidth]{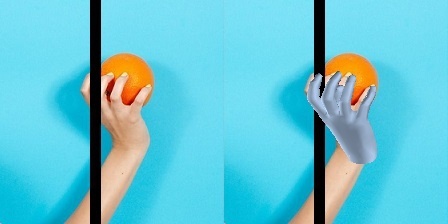}
\includegraphics[trim=0 0 0 0, clip,width=0.33\textwidth]{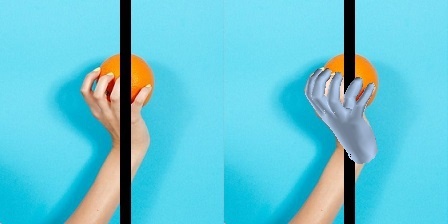}
\includegraphics[trim=0 0 0 0, clip,width=0.33\textwidth]{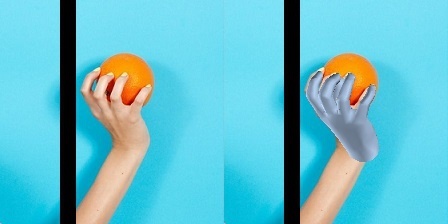}
\includegraphics[trim=0 0 0 0, clip,width=0.33\textwidth]{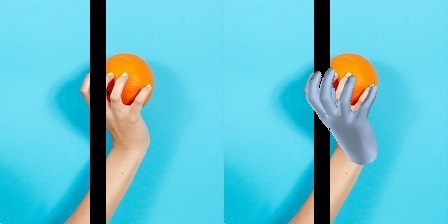}
\includegraphics[trim=0 0 0 0, clip,width=0.33\textwidth]{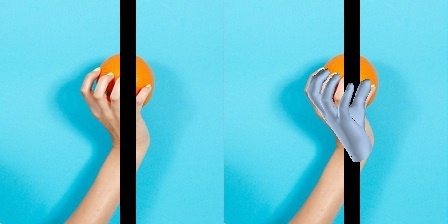}
\includegraphics[trim=0 0 0 0, clip,width=0.33\textwidth]{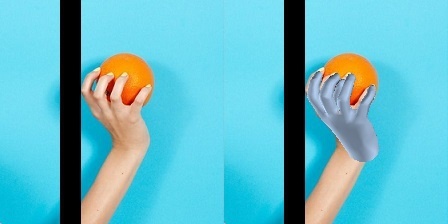}
\includegraphics[trim=0 0 0 0, clip,width=0.33\textwidth]{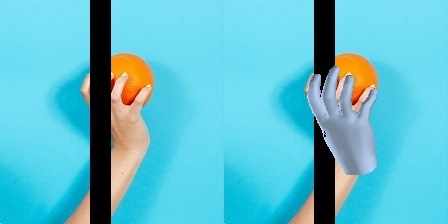}
\includegraphics[trim=0 0 0 0, clip,width=0.33\textwidth]{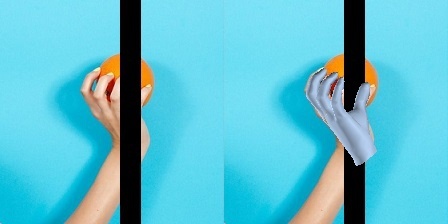}
\vspace{-2mm}
\setlength{\tabcolsep}{30pt}
\begin{tabular}{cccccc}
Input & Output &  Input & Output &  Input & Output
\end{tabular}
\caption{
Qualitative results of our method. There is a hand with an orange. We demonstrate the robustness of our model by adding artificial occlusions including black vertical stripes to the images. We can see that Graphormer reconstructs plausible hand mesh under the occlusion scenarios. Please see \href{https://github.com/microsoft/MeshGraphormer/blob/main/docs/Fig1.gif}{\texttt{Fig1.gif}} for more detailed video results.
} 
\label{fig:vis_hand_1}
\end{center}
\end{figure*}

\begin{figure*}
\begin{center}
\includegraphics[trim=0 0 0 0, clip,width=0.33\textwidth]{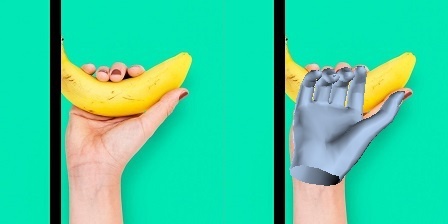}
\includegraphics[trim=0 0 0 0, clip,width=0.33\textwidth]{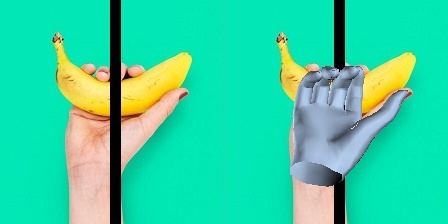}
\includegraphics[trim=0 0 0 0, clip,width=0.33\textwidth]{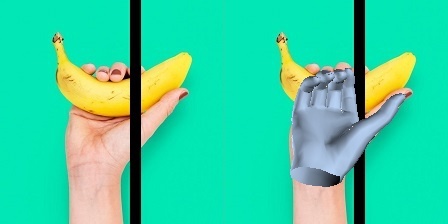}
\includegraphics[trim=0 0 0 0, clip,width=0.33\textwidth]{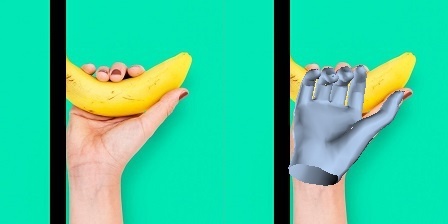}
\includegraphics[trim=0 0 0 0, clip,width=0.33\textwidth]{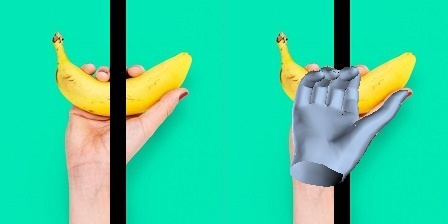}
\includegraphics[trim=0 0 0 0, clip,width=0.33\textwidth]{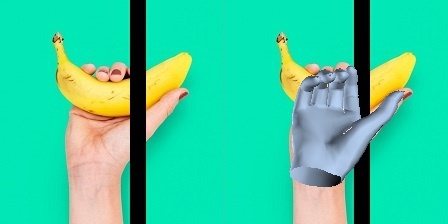}
\includegraphics[trim=0 0 0 0, clip,width=0.33\textwidth]{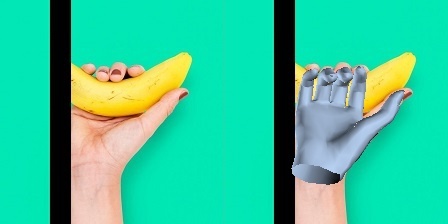}
\includegraphics[trim=0 0 0 0, clip,width=0.33\textwidth]{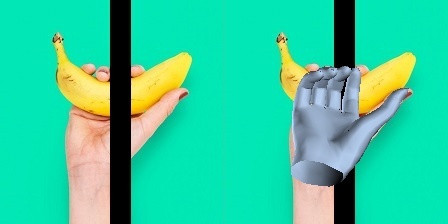}
\includegraphics[trim=0 0 0 0, clip,width=0.33\textwidth]{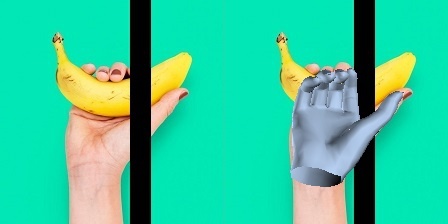}
\vspace{-2mm}
\setlength{\tabcolsep}{30pt}
\begin{tabular}{cccccc}
Input & Output &  Input & Output &  Input & Output
\end{tabular}
\caption{
Qualitative results of our method. There is a hand holding a banana. We do not have any banana training images. However, Graphormer generalizes to the novel object, and creates the hand mesh with the correct pose. Please see \href{https://github.com/microsoft/MeshGraphormer/blob/main/docs/Fig2.gif}{\texttt{Fig2.gif}} for more detailed video results.
} 
\label{fig:vis_hand_2}
\end{center}
\end{figure*} 

\begin{figure*}
\begin{center}
\includegraphics[trim=0 0 0 0, clip,width=0.33\textwidth]{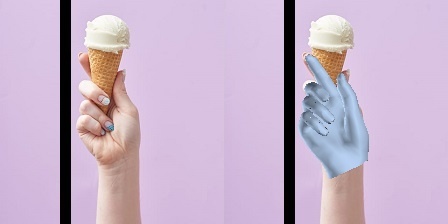}
\includegraphics[trim=0 0 0 0, clip,width=0.33\textwidth]{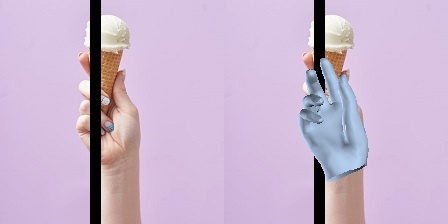}
\includegraphics[trim=0 0 0 0, clip,width=0.33\textwidth]{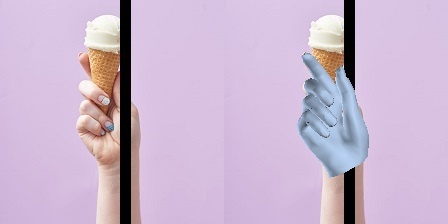}
\includegraphics[trim=0 0 0 0, clip,width=0.33\textwidth]{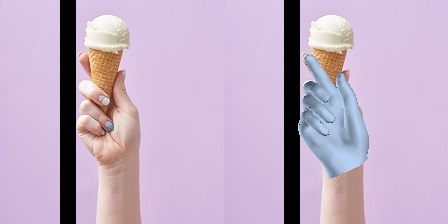}
\includegraphics[trim=0 0 0 0, clip,width=0.33\textwidth]{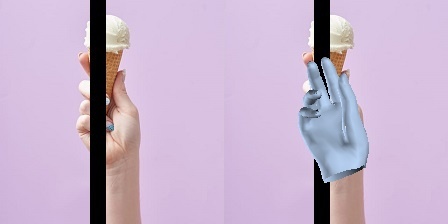}
\includegraphics[trim=0 0 0 0, clip,width=0.33\textwidth]{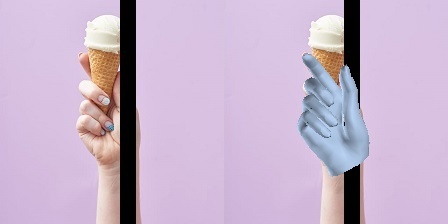}
\includegraphics[trim=0 0 0 0, clip,width=0.33\textwidth]{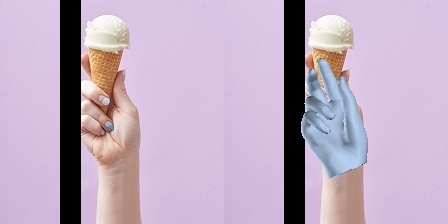}
\includegraphics[trim=0 0 0 0, clip,width=0.33\textwidth]{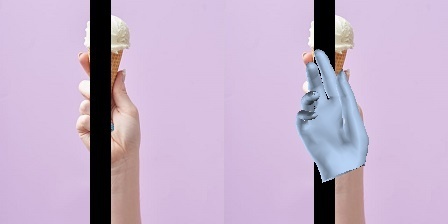}
\includegraphics[trim=0 0 0 0, clip,width=0.33\textwidth]{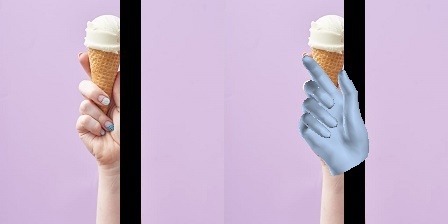}
\vspace{-2mm}
\setlength{\tabcolsep}{30pt}
\begin{tabular}{cccccc}
Input & Output &  Input & Output &  Input & Output
\end{tabular}
\caption{
Qualitative results of our method. There is a hand holding an ice cream cone. The ice cream cone is a novel object unseen in training, and the hand pose is also object-specific. We can see that Graphormer works reasonably well for the test images. Please see  \href{https://github.com/microsoft/MeshGraphormer/blob/main/docs/Fig3.gif}{\texttt{Fig3.gif}} for more detailed video results.
} 
\label{fig:vis_hand_3}
\vspace{-2mm}
\end{center}
\end{figure*} 

\begin{table*}
\centering
\begin{tabular}{lccc}
    \toprule
	Method  & Para (M) & $\Delta$ Para (M) & PAMPJPE $\downarrow$ \\
	\midrule
	Graphormer - GRB & $98.39$ & $-$ & $35.9$\\
	Graphormer - GRB + MLP1 & $98.43$ & $0.04$ & $35.9$\\
    Graphormer - GRB + MLP2 & $98.92$ & $0.53$ & $36.0$\\
	\midrule
	Graphormer & $98.43$ & $0.04$ & $34.5$\\
	\bottomrule
\end{tabular}
\caption{Comparison between the use of MLP and GRB.}
\label{tbl:mlp}
\vspace{-2mm}
\end{table*}

\begin{figure*}
\begin{center}
\includegraphics[trim=0 0 0 0, clip,width=0.33\textwidth]{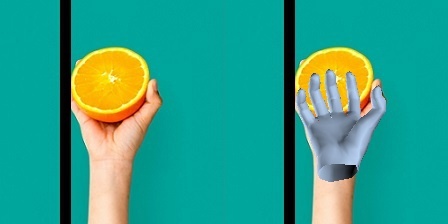}
\includegraphics[trim=0 0 0 0, clip,width=0.33\textwidth]{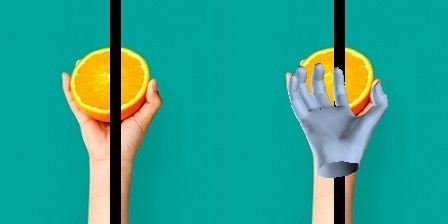}
\includegraphics[trim=0 0 0 0, clip,width=0.33\textwidth]{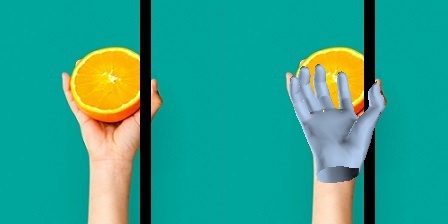}
\includegraphics[trim=0 0 0 0, clip,width=0.33\textwidth]{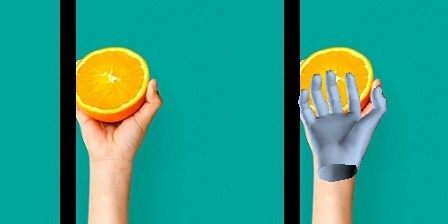}
\includegraphics[trim=0 0 0 0, clip,width=0.33\textwidth]{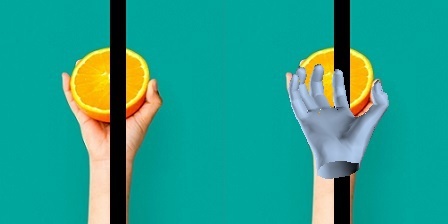}
\includegraphics[trim=0 0 0 0, clip,width=0.33\textwidth]{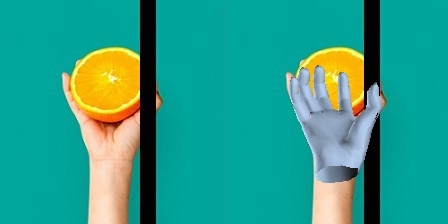}
\includegraphics[trim=0 0 0 0, clip,width=0.33\textwidth]{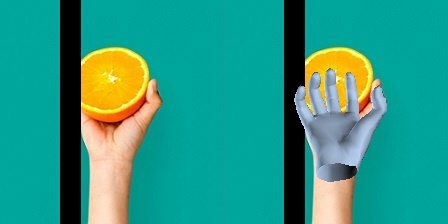}
\includegraphics[trim=0 0 0 0, clip,width=0.33\textwidth]{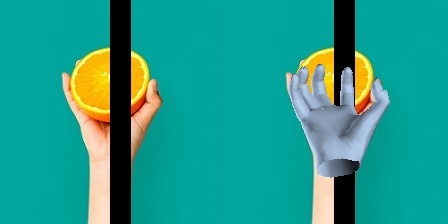}
\includegraphics[trim=0 0 0 0, clip,width=0.33\textwidth]{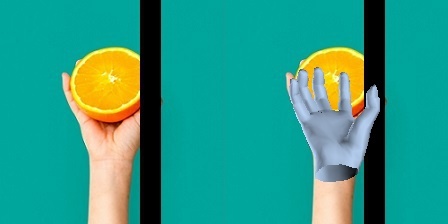}
\vspace{-2mm}
\setlength{\tabcolsep}{30pt}
\begin{tabular}{cccccc}
Input & Output &  Input & Output &  Input & Output
\end{tabular}
\caption{
Qualitative results of our method. It is a hand holding half an orange. Graphormer is able to reconstruct a reasonable hand mesh even though most of the fingers are invisible. Please see \href{https://github.com/microsoft/MeshGraphormer/blob/main/docs/Fig4.gif}{\texttt{Fig4.gif}} for more detailed video results.
} 
\label{fig:vis_hand_4}
\end{center}
\end{figure*}

\section{\camready{Qualitative Comparison}}

\camready{Figure~\ref{fig:occlusions} shows qualitative results of Graphormer compared with METRO~\cite{lin2020end} in the scenario of heavy occlusions. We can see that both methods are quite robust against occlusions, but Graphormer generates better head and body poses. At the top right of Figure~\ref{fig:occlusions}, almost half of the subject is occluded. Graphormer reconstructs a human mesh with more accurate head/body pose compared to METRO. At the bottom right of Figure~\ref{fig:occlusions}, the subject is occluded by the car door. We see Graphormer reconstructs a more reasonable body shape. At the bottom left, the subject is standing behind the fence. Our method reconstructs a human mesh with the two legs better aligned with the image. The results demonstrate the effectiveness of the proposed method.}

\section{Additional Qualitative Results}

Further, to demonstrate the robustness and generalization capability of our model to challenging scenarios, we test our model on the hand images that are collected from the Internet. The images have severe occlusions with different objects. 

To make the task even more difficult, we create artificial occlusions including black vertical stripes to cover one or two fingers, or part of the palm of the hand in the test images. Please note that the artificial occlusions are only used in the inference stage. We do not use any artificial occlusions in training.

In Figure~\ref{fig:vis_hand_1}, Figure~\ref{fig:vis_hand_2}, Figure~\ref{fig:vis_hand_3}, and Figure~\ref{fig:vis_hand_4}, we show the input images and our reconstructed hand meshes. For each figure, the top row shows the occlusion scenario with narrow black stripes. From the second row to the bottom, we gradually increase the width of the stripes to occlude more fingers or more parts of the hand. 

Figure~\ref{fig:vis_hand_1} shows a hand with an orange. Our model is able to reconstruct a reasonable hand mesh, even if the hand is severely obscured by the vertical stripes. The results show that Graphormer is to some extent robust to the artificial occlusion patterns.

In Figure~\ref{fig:vis_hand_2}, there is a hand grasping a banana. Although the banana is a novel object unseen in training and a large portion of the fingers are occluded by the banana, Graphormer successfully reconstructs the hand mesh under various occlusion scenarios. This demonstrates the generalization ability of our proposed Graphormer.

Figure~\ref{fig:vis_hand_3} shows a hand with an ice cream cone. Please note that the ice cream cone is unseen during training, and the interaction between the hand and the ice cream cone is complex. However, our model generalizes well to the test images. Even though the occlusions by the ice cream cone are severe and sometimes most of the fingers are occluded by the black vertical stripes, Graphormer still reconstructs a reasonable hand shape with object-specific grasp. 

Figure~\ref{fig:vis_hand_4} shows a hand with half an orange. Most of the fingers are invisible in this image. However, our model creates a hand mesh with the correct hand pose.

We further present the video results of Figure~\ref{fig:vis_hand_1}, Figure~\ref{fig:vis_hand_2}, Figure~\ref{fig:vis_hand_3}, and Figure~\ref{fig:vis_hand_4}. Please find the video results in the attached GIF files.

\section{\camready{Comparison between MLP and GCN}}

\camready{In this section, we replace graph residual block with MLPs, and study the performance of the use of MLPs with a similar or larger model size.}

\camready{In Table~\ref{tbl:mlp}, the first row corresponds to the baseline transformer that uses image grid features. In the second and the third rows, we gradually increase the hidden size of the MLP module in the transformer, but we do not achieve any gain in performance. The bottom row of Table~\ref{tbl:mlp} shows the results of our Graphormer. As can be seen, adding graph convolutions has a slight increase of 0.04M parameters, but it improves performance significantly from 35.9 to 34.5 PA-MPJPE.} 

\begin{figure}
\begin{center}
\includegraphics[trim=0 0 0 0, clip,width=0.9\columnwidth]{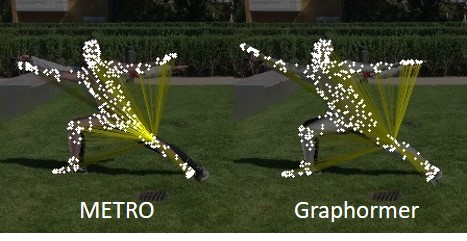}
\caption{
Attention map without color normalization. Graphormer pays more attentions to the lower left leg compared to METRO.} 
\label{fig:att}
\vspace{-2mm}
\end{center}
\end{figure}

\begin{figure*}[t]
\begin{center}
\includegraphics[trim=0 0 0 0, clip,width=0.8\textwidth]{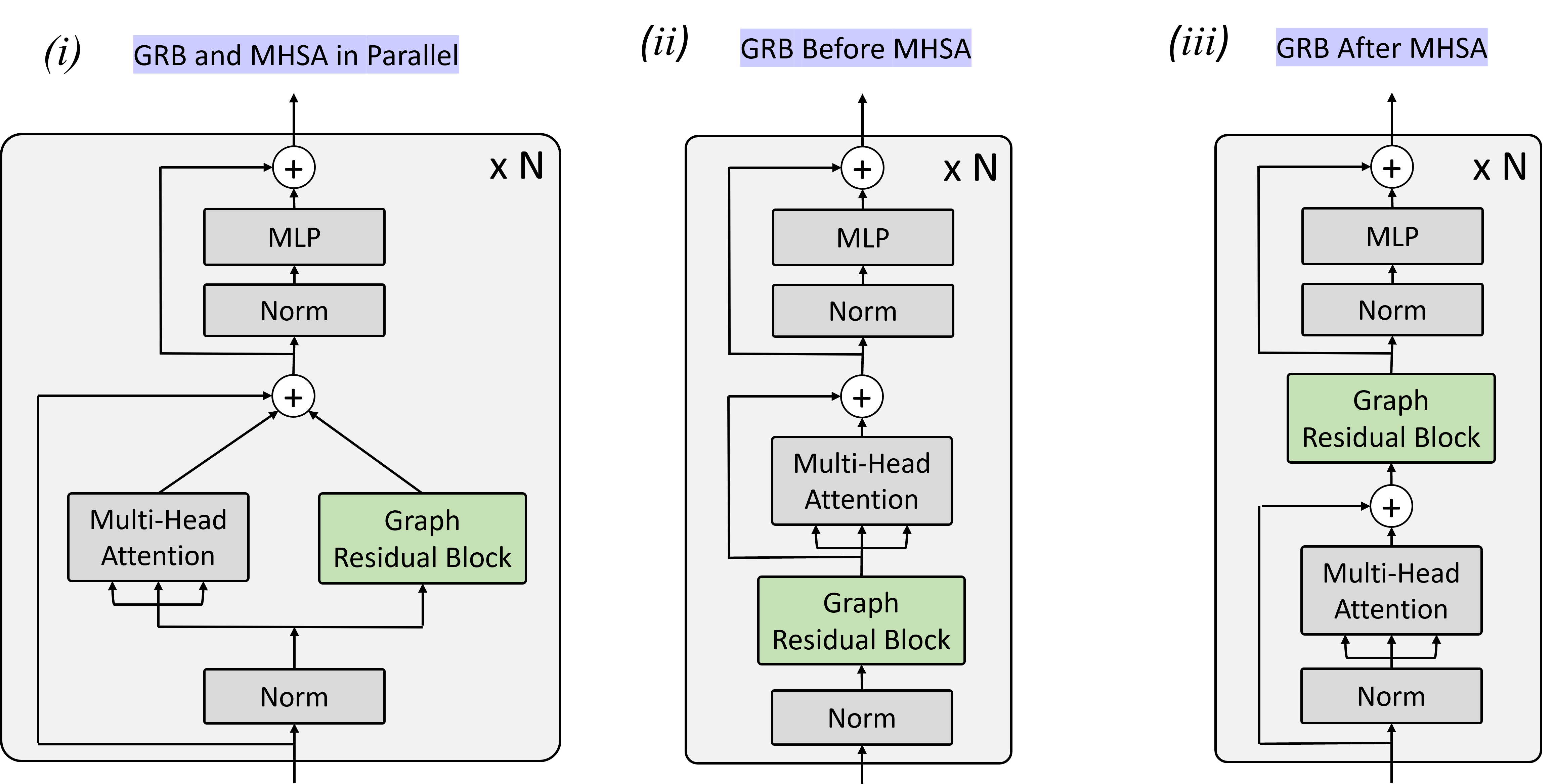}
\caption{
Three design options we have studied for building our proposed Graphormer Encoder. The designs are inspired by language and speech literature~\cite{gulati2020conformer,wu2020lite}.} 
\label{fig:different_choices}
\vspace{-2mm}
\end{center}
\end{figure*}

\section{\camready{Design Options of Graphormer Encoder}}

\camready{In Figure~\ref{fig:different_choices}, we graphically illustrate three design options of the Graphormer encoder we have studied in the paper. Please refer to Table 6 in our main paper for the performance comparisons between the design options. We observe that placing graph convolutions after MHSA works better than other design options for the reconstruction of human mesh.}

\section{\camready{Discussion of Attention Map}}
\camready{Please note that the attention colors in the paper's diagrams are normalized based on the maximum attention value. Because the maximum attention value for Graphormer is smaller, so the overall colors are lighter. We attach the two diagrams without color normalization in Figure~\ref{fig:att}. We see that both methods pay similar attention on the left arm and right foot, while Graphormer also attends to the left lower leg.} 

\section{\camready{Discussion of Camera Parameters}}

\camready{We learn camera parameters for a weak perspective camera model. Following~\cite{kolotouros2019convolutional}, we predict a scaling factor $s$ and a 2D translation vector $t$. Please note that the model prediction is already in the camera coordinate system, thus we don't have to compute global camera rotation. The camera parameters are learned via 2D pose re-projection optimization. It doesn't require any GT camera parameters.}

\begin{table}
\centering
\begin{tabular}{lccc}
    \toprule
	 & HRNet & Transformer & Graphormer\\
	\midrule
	
	\# Parameters (M) & $128.05$ & $98.39$ & $98.43$ \\ 
	GFLOPs & $28.89$ & $27.71$ & $27.72$ \\ 
	\bottomrule
\end{tabular}
\caption{Number of parameters and computational complexity in terms of GFLOPs.}
\label{tbl:complexity}
\end{table}

\section{\camready{Training Time}}
\camready{We conducted experiments on a machine with 8 NVIDIA V100 GPUs. We use a batch size of 32. For each epoch, our training takes about 35 minutes. We train the proposed model for 200 epochs. The overall training takes 5 days.}

\section{Computational Costs}
Since we inject graph convolutions into the transformer, one may wonder about the computational costs of the proposed Graphormer. To answer the question, we report the number of parameters and the computational complexity in terms of GFLOPs. 

Table~\ref{tbl:complexity} shows the comparison between the conventional transformer and the proposed Graphormer. We also report the computational cost of the HRNet CNN backbone for reference. As we can see, adding graph convolutions has a slight increase of $0.04$M parameters and $0.01$ GFLOPs compared to the conventional transformer. The results suggest that little complexity has been added to the transformer architecture. However, Graphormer significantly improves the state-of-the-art performance across multiple benchmarks. This verifies the effectiveness of the proposed method.

\camready{Please note that the total parameters of our end-to-end pipeline is the sum of HRNet and Graphormer.}

\section{\camready{Limitation}}
\begin{figure}
\begin{center}
(a) Example1
\includegraphics[width=1.\columnwidth]{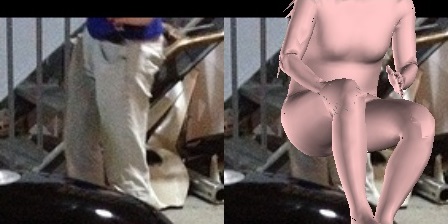}\\
\setlength{\tabcolsep}{45.0pt}
\begin{tabular}{cc}
Input & Output\\
\end{tabular}
(b) Example2\\
\includegraphics[width=1.\columnwidth]{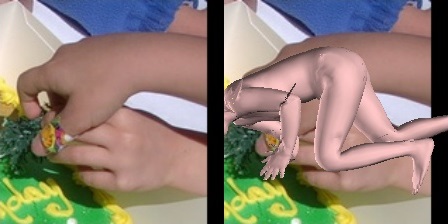}\\
\setlength{\tabcolsep}{45.0pt}
\begin{tabular}{cc}
Input & Output\\
\end{tabular}
\caption{
Failure cases. Mesh Graphormer may not work well if the reconstruction target is out of the view.
} 
\vspace{-0mm}
\label{fig:limitation}
\end{center}
\end{figure}

\camready{We observed that our method may not work well if the reconstruction target is out of the view. For example, as shown in Figure~\ref{fig:limitation}(a), when the majority of the human body is not in the input image, our method fails to estimate a correct human mesh. This is probably due to the lack of out-of-the-view 3D training data in our training set. In Figure~\ref{fig:limitation}(b), only two hands are visible and the rest of the human body is out of the view. Our method does not work well in this case. We plan to address this issue in our future work.
}

\end{document}